\documentclass{article}

\usepackage{microtype}
\usepackage{graphicx}
\usepackage{subfigure}
\usepackage{booktabs} 

\usepackage{amsmath}
\usepackage{multicol,multienum, multirow}
\usepackage{algorithmic}
\usepackage{algorithm}
\usepackage{booktabs}
\usepackage{xcolor}
\usepackage{threeparttable}
\usepackage{natbib}
\usepackage{caption}

\usepackage{hyperref}

\usepackage[accepted]{icml2021}

\icmltitlerunning{ }

\begin{document}

\twocolumn[
\icmltitle{Multi-Model Least Squares-Based\\ Recomputation Framework for Large Data Analysis}



\icmlsetsymbol{equal}{*}

\begin{icmlauthorlist}
\icmlauthor{Wandong Zhang}{Win,Lak}
\icmlauthor{QM Jonathan Wu}{Win}
\icmlauthor{Yimin Yang}{Lak,VI}
\icmlauthor{WG Will Zhao}{Lak2,Qu}
\icmlauthor{TianLei Wang}{S1}
\icmlauthor{Hui Zhang}{Hu}
\end{icmlauthorlist}

\icmlaffiliation{Win}{Department of Electrical and Computer Engineering, University of Windsor, Windsor, Canada}
\icmlaffiliation{Lak}{Department of Computer Science, Lakehead University, Thunder Bay, Canada}
\icmlaffiliation{VI}{Vector Institute for Artificial Intelligence, Toronto, Canada}
\icmlaffiliation{Lak2}{Faculty of Business Administration, Lakehead University}
\icmlaffiliation{Qu}{Artificial Intelligence Group, CEGEP de Ste Foy, Canada}
\icmlaffiliation{S1}{Institute of Information and Control, Hangzhou Dianzi University, Hangzhou, China}
\icmlaffiliation{Hu}{School of Robotics, Hunan University, Changsha, China}

\icmlcorrespondingauthor{QM Jonathan Wu}{jwu@uwindsor.ca}

\icmlkeywords{Machine Learning, ICML}

\vskip 0.3in
]



\printAffiliationsAndNotice{\icmlEqualContribution} 

\begin{abstract}

Most multilayer least squares (LS)-based neural networks, such as deep random vector functional link (RVFL), are structured with two separate stages: unsupervised feature encoding and supervised pattern classification. Once the unsupervised learning is finished, the latent encoding would be fixed without supervised fine-tuning. However, in complex tasks such as handling the ImageNet dataset, there are often many more clues that can be directly encoded, while the unsupervised learning, by definition, cannot know exactly what is useful for a certain task. There is a need to retrain the latent space representations in the supervised classification stage to learn some clues that unsupervised learning has not yet learned. In particular, the error matrix from the output layer is pulled back to each hidden layer, and the parameters of the hidden layer are recalculated with Moore-Penrose (MP) inverse for more generalized representations. In this paper, a recomputation-based multilayer network using MP inverse (RML-MP) is developed. A sparse RML-MP (SRML-MP) model to boost the performance of RML-MP is then proposed. The experimental results with varying training samples (from $3\,K$ to $1.8\,M$) show that the proposed models provide better generalization performance than most representation learning algorithms.


\end{abstract}


\section{Introduction}
\label{submission}

As technology improves, the demand for high-dimensional big-data analysis has increased tremendously. Datasets with many samples, though can provide greater statistical power, are prone to higher complexity and redundancy. To handle the high-dimensional features, much effort has been dedicated to developing representation learning (RL) algorithms that reduce irrelevant details while preserving the discriminative information beneficial for final pattern recognition~\cite{bengio2013representation}. If the represented features are compact and relatively low dimensional, classifying the unknown high-dimensional patterns would not be difficult~\cite{chu2020unsupervised}.

The existing RL algorithms can be divided into three categories: the probabilistic-based algorithms~\cite{bengio2013representation}, the geometrically motivated manifold learning models~\cite{maaten2008visualizing}, and the reconstruction-based approaches~\cite{bengio2007greedy}. The most common architecture of reconstruction-based approaches is that of the auto-encoders (AEs). The AE is an unsupervised learning algorithm that generally applies an iterative learning strategy such as backpropagation (BP) as a cornerstone of its training, aiming to learn the reduced encoding by reproducing the input patterns at the output layer. The AE was first introduced in~\cite{ballard1987modular} as a way of pretraining in ANNs. Then, the structure was employed for representation learning, such as sparse AE (SAE), denoise AE (DAE), and weight-decay regularization-based AE (WD-AE).

Going beyond simply learning with iterative learning strategies, research in the last decade has paid attention to AEs trained with non-iterative LS-based MP inverse techniques. Compared to the BP learning scheme that may become trapped in a local minimum and is sensitive to the learning rate setting, the LS-based strategies have the advantage of quicker convergence. The earliest attempt of the MP inverse technique in a neural network can be traced back to 1992~\cite{schmidt1992feedforward}. Recently, the random thoughts penetrate neural networks, several LS-based algorithms, such as RVFL~\cite{igelnik1995stochastic}, have been proposed. The RVFL, which adopts random hidden layer neurons (generated within a suitable range and kept fixed) and generates the output weights with the MP inverse, is a single layer neural network (SLNN) with a fast training speed and excellent generalization performance. Based on that, the LS-based AE~\cite{kasun2013representational, katuwal2019stacked} using MP inverse for unsupervised encoding has been proposed. In recent years, explosive developments on multilayer RL networks using MP inverse have been seen~\cite{huang2019stochastic, zhang2019unsupervised}. These networks have key characteristics in common: i) they stack LS-based AEs to capture the latent space representations, and ii) these models show state-of-the-art performance on small-scale tasks.



Nonetheless, though the existing LS-based RL algorithms show remarkable performance in image classification and pattern recognition, they cannot be directly applied in big data analysis due to several limitations: First, they cannot obtain satisfactory results on high-dimensional datasets with a large number of training samples. Most LS-based RL algorithms have focused on processing small-scale and medium-scale datasets, such as MNIST and NORB with no more than $100 K$ samples, whereas utilizing MP inverse to handle big datasets, such as ImageNet containing more than 1.2 million patterns and Place365 with a sample size of more than 1.8 million, has rarely been explored.

Second, most of the MP inverse-based RL frameworks generate loosely connected representations in processing big datasets. The LS-based AE belongs to the unsupervised dimension reduction since no labels are included. In most cases, LS-based AE is an intermediate model toward the final objects. For example, Katuwal \textit{et al.}~\cite{katuwal2019random} proposed an MP inverse-based deep network using multiple LS-based AEs for feature extraction. Then, the weights of the final layer have been analytically calculated on the low-dimensional representation using regularized LS. In contrast to BP-based AEs for which the parameters of each AE would be fine-tuned according to the label information, all the state-of-the-art multilayer LS-based networks do not contain fine-tuning~\cite{wang2020hierarchical}. Therefore, the parameters of LS-based AEs are fixed once the feature is extracted. Such a training process makes LS-based multilayer networks achieve fast training speed and comparable performance on small-scale datasets over some BP-based deep networks. However, lacking supervision, some important clues may be filtered before training the final layer, which affects the final performance~\cite{chao2019recent}. This issue is more obvious when handling large-scale sets with high complexity.

\begin{figure}[!t]
\centering
\includegraphics[trim={1.0cm, 1.0cm, 1.0cm, 1.0cm}, clip, width=0.5\textwidth]{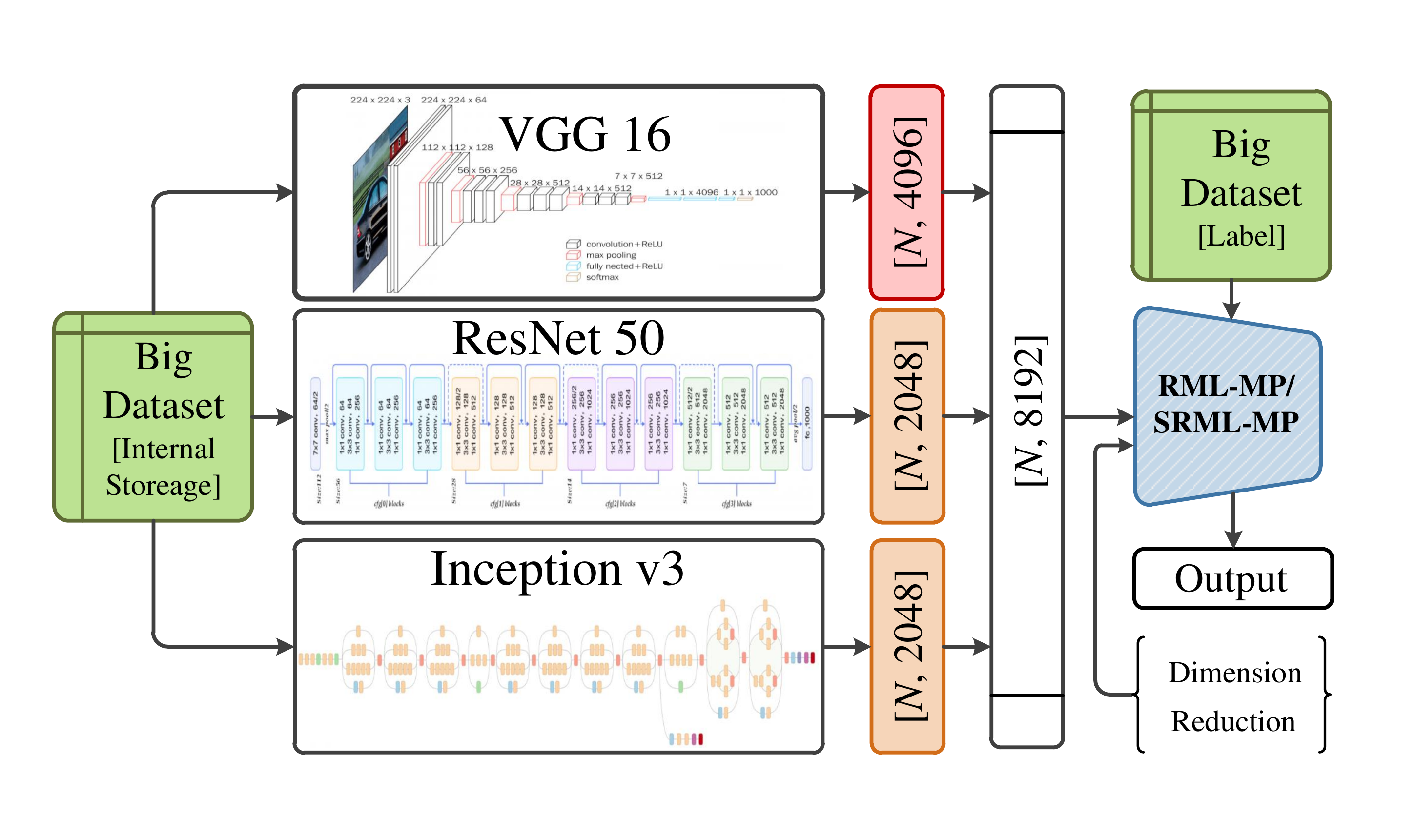}
\vspace{-0.8 cm}
\caption{Flowchart of the RML-MP and SRML-MP for multi-model representation learning.}
\vspace{-0.3 cm}
\label{a1}
\end{figure}

To address the above-mentioned limitations, this paper proposes a novel LS-based algorithm called RML-MP for big data analysis. First, the LS-based AEs using $\ell_2$ penalty are stacked to exploit the effective encoding from complex input data, and the final classification layer is calculated with an LS scheme. Then, the LS strategy is applied to pull back the current error from the output layer to each hidden layer one-by-one, generating the desired output $\textit{\textbf P}$ for each hidden layer. Finally, based on the input data and desired output, the MP inverse technique is utilized to recompute weights in each layer. By doing so, the bond between hidden layer representations and labels is strengthened, and robust representations can be obtained. Meanwhile, the effective $\ell_{1/2}$ penalty-based learning framework is adopted in the retraining of RML-MP, leading to a sparse algorithm SRML-MP. The $\ell_{1/2}$ penalty in retraining helps SRML-MP to obtain sparse weights. The structures are depicted in Fig.~\ref{a1}.

Furthermore, as stated in~\cite{zhang2015ensemble, wang2015unsupervised}, for most real applications, multi-model feature learning is more practical than single-model data learning. It is thus of great interest to exploit multi-model big data learning for higher performance purposes. This paper makes three contributions to the extant knowledge. \textcolor{black}{First, an RML-MP framework is proposed to perform multi-model big data encoding.} Second, an SRML-MP is further developed to improve the performance of RML-MP. Third, a comprehensive comparison is conducted to validate the effectiveness of the proposed models over RL approaches on different datasets, such as ImageNet and Place365.

\section{Methodology}
\label{meth}

\begin{figure*}[!t]
\begin{minipage}[b]{1\linewidth}
\centering
\includegraphics[trim={0.8cm, 0.5cm, 1.2cm, 1.0cm}, clip, width=\textwidth]{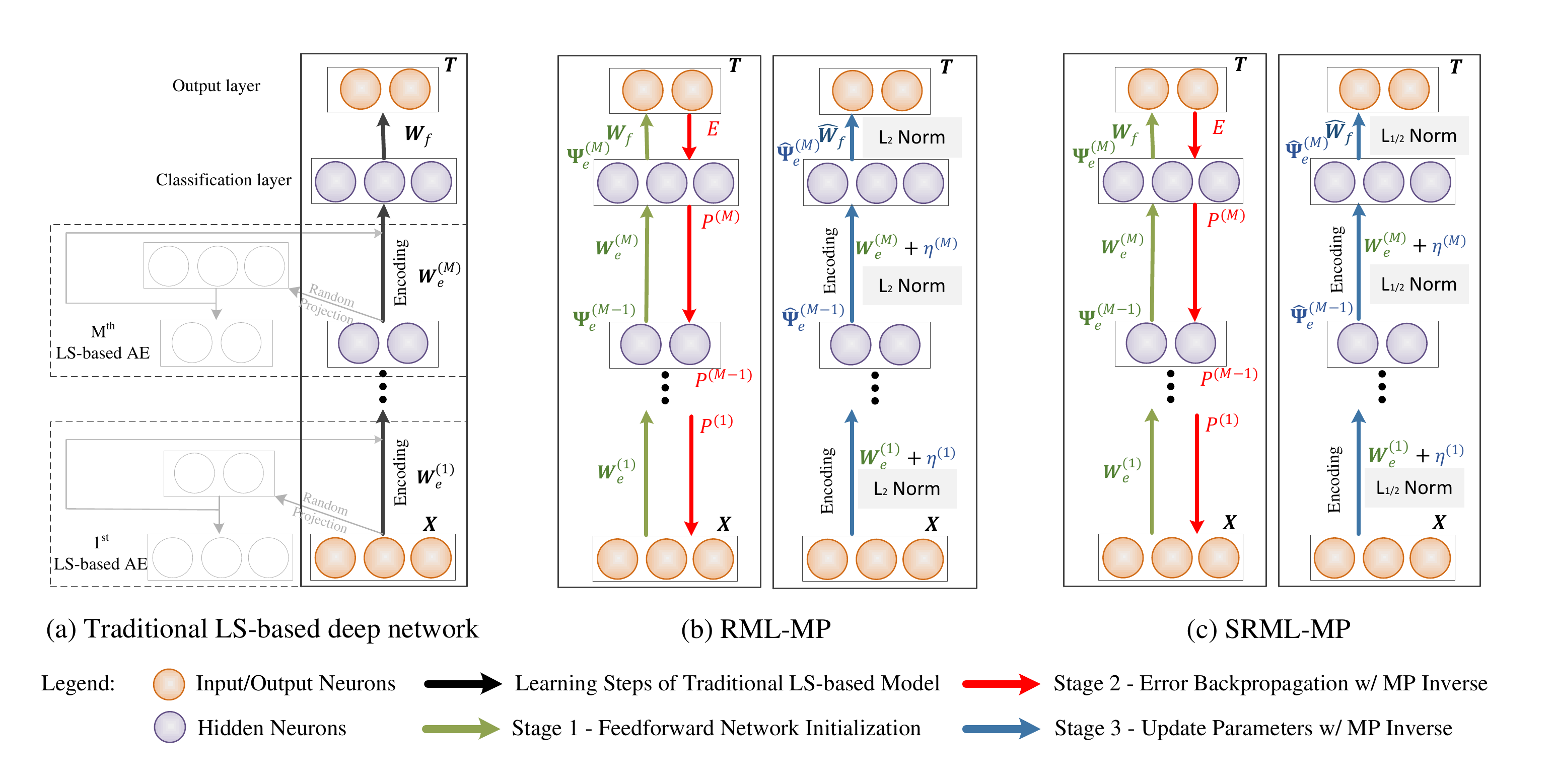}
\end{minipage}
\vspace{-0.8cm}
\caption{Comparison of frameworks of the (a) traditional LS-based deep network~\cite{katuwal2019stacked}, (b) the proposed RML-MP, and (c) the proposed SRML-MP. The difference of RML-MP and SRML-MP lies in Stage 3, the RML-MP use $\ell_2$ penalty to recalculate the parameters while the SRML-MP adopt $\ell_{1/2}$ penalty to update the weights.}
\label{asa}
\vspace{-0.2cm}
\end{figure*}

Figure~\ref{asa} shows the diagrams of the traditional LS-based deep model, the proposed RML-MP, and the SRML-MP algorithms for comparison purposes.

\subsection{The Proposed RML-MP}

Table~\ref{define} describes the notations used in this paper. The main difference of the proposed models and other traditional LS-based neural networks~\cite{katuwal2019stacked, katuwal2019random} lies in the model training procedures: The proposed RML-MP contains three successive learning stages: Stage 1 - the feedforward initialization, Stage 2 - backpropagation with MP inverse, and Stage 3 - update weights with MP inverse.

\begin{table}[!t]
\small
\vspace{-0.3 cm}
\caption{Notations to be used in this paper}
\vspace{0.2 cm}
\begin{tabular}{cl}
\toprule
Notation   & Meaning     \\
\midrule
$C$                 & regularization term $C$\\
$M$                 & the number of AEs in RML-MP / SRML-MP\\
$\lambda$           & learning rate\\
$\textbf{\textit{T}}$                 & the target output\\
$\textbf{\textit{Y}}$                 & the network output\\
$\textbf{\textit{W}}_e^{(i)}$           & parameters of the $i$-{th} encoding layer\\
$\hat{\textbf{\textit{W}}}_e^{(i)}$     & updated parameters of the $i$-{th} encoding layer\\
$\pmb{\eta}^{(i)}$              & offset weights of the $i$-{th} encoding layer\\
$\Psi_e^{(i)}$             & the encoding of the $i$-{th} layer\\
$\hat{\Psi}_e^{(i)}$       & the updated encoding of the $i$-{th} layer\\
$\textbf{\textit{W}}_f$               & parameters of the last classification layer\\
$\hat{\textbf{\textit{W}}}_f$         & updated parameters of the last classification layer\\
$\textbf{\textit{E}}$			        & error matrix from the output layer\\
$\textbf{\textit{P}}^{(i)}$               & the $i$-{th} layer error feedback data\\
\bottomrule
\vspace{-0.2 cm}
\label{define}
\end{tabular}
\end{table}

\subsubsection{Stage 1: Feedforward Initialization} The first stage aims to develop a traditional LS-based deep network. Given a training dataset with $N$ number of samples, $\textbf{\textit{X}}=\{\textbf{\textit{X}}_1, \textbf{\textit{X}}_2, \cdots, \textbf{\textit{X}}_N\}$, $\textbf{\textit{X}}_i\in \mathcal R^{n}$ is the input vector, $\textbf{\textit{T}}=\{\textbf{\textit{T}}_1, \textbf{\textit{T}}_2, \cdots, \textbf{\textit{T}}_N\}\in \mathcal R^d$ is its associated output target.

The LS-based AE~\cite{katuwal2019stacked, katuwal2019random, kasun2013representational} tries to encode the input data through setting the input as the target output. In particular, the optimal output layer parameters of LS-based AE are calculated with the MP inverse technique. With randomly assigned hidden layer weights $\textbf{\textit{W}}$, the $\ell_2$ norm-based AE is optimized with the following minimizing problem:
\begin{equation}
\small
\label{n1}
\begin{split}
&{\rm min}\,\,J=\frac{1}{2}||\Psi_e\textbf{\textit{W}}_e-\textbf{\textit{X}}||^2_F+\frac{C}{2}||\textbf{\textit{W}}_e||^2_F,\\
&{\rm s. t. }\,\,\Psi=\sigma(\textbf{\textit{X}},\textbf{\textit{W}}), \,\,{\rm and}\\
&\,\,\,\,\,\,\,\,\,\,\textbf{\textit{W}}^T\textbf{\textit{W}}=I,\\
\end{split}
\end{equation}
where $\sigma(\cdot)$ is the activation function (sine or sigmoid), $C$ is the regularization term, $\Psi_e$ is the hidden space encoding, and $\textbf{\textit{W}}_e$ refers to the output layer weight, which is calculated with MP inverse: $\textbf{\textit{W}}_{e}=\Psi_e^\dagger\textit{\textbf{X}}$. In this paper, with the identity matrix $I$, the LS fit is utilized:
\begin{equation}
\small
\textbf{\textit{W}}_{e}=\Psi_e^\dagger\textit{\textbf{X}}=(\frac{I}{C}+\Psi_e^T\Psi_e)^{-1}\Psi_e^T\textit{\textbf{X}}.
\end{equation}

The encoding $\Psi_{e}$ of the LS-based AE is described by
\begin{equation}
\small
\Psi_{e}=\sigma\left(\textbf{\textit{X}}\cdot \textbf{\textit{W}}_e^T\right).
\end{equation}

Supposing that a multilayer LS-based network has $M$ hidden layers.  Mathematically, the objective function $J$ to learn this deep network can be described as:
\begin{equation}
\small
\label{n2}
\begin{split}
&{\rm min}\,\,J=\frac{1}{2}||\Psi_{e}^{(M) }\textbf{\textit{W}}_f-\textbf{\textit{T}}||^2_F+\frac{C}{2}||\textbf{\textit{W}}_f||^2_F,\\
&{\rm s. t. }\,\,\Psi^{(i)}_e=\sigma\left(\Psi^{(i-1)}_e\cdot \left(\textbf{\textit{W}}_e^{(i-1)}\right)^T\right), 1\leq i\leq M,\\
\end{split}
\end{equation}
where $\Psi^{(i)}_e$ is the $i$-{th} layer output matrix, $\textbf{\textit{X}}$ can be considered as the $0$-{th} layer feature matrix ($\Psi^{(i)}_e$ where $i$ equal to zero), $\textbf{\textit{W}}_f$ refers to the weights of the classification layer which is generated through MP inverse, and $\textbf{\textit{W}}_e^{(i)}$ is the $i$-{th} layer weights that would be fixed once determined.

The output layer weights $\textbf{\textit{W}}_f$ are calculated as Eq.~(\ref{n3}).
\begin{equation}
\small
\label{n3}
\begin{split}
&\textbf{\textit{W}}_f=\left(\frac{I}{C}+(\Psi_e^{(M)})^T\Psi_e^{(M)}\right)^{-1}(\Psi_e^{(M)})^T\textbf{\textit{T}},\\
&\textbf{\textit{Y}}=\Psi_e^{(M)}\cdot \textbf{\textit{W}}_f.\\
\end{split}
\end{equation}

\subsubsection{{Stage 2: Backpropagation with MP Inverse}} For the last classification layer, the target output $\textbf{\textit{T}}$ and the network output $\textbf{\textit{Y}}$ are obtained. Next, the network error $\textbf{\textit{E}}$ is pulled back from the output layer to each hidden layer. In fact, the proposed retraining strategy mainly aims to adjust the hidden layer representations by attempting to offset the pulled error term. Specifically, the error of the output layer can be described as
\begin{equation}
\small
\label{n4}
\textbf{\textit{E}}=\textbf{\textit{T}}-\Psi_e^{(M)}\cdot \textbf{\textit{W}}_f.\\
\end{equation}

First, we pull the error back across the last classification layer, the desired change of the last hidden layer according to $\textbf{\textit{E}}$ and current $\textbf{\textit{W}}_f$ is
\begin{equation}
\small
\label{n5}
\textbf{\textit{P}}^{(M)}=\textbf{\textit{E}} \cdot \left(\frac{I}{C}+{\textbf{\textit{W}}_f}^T\textbf{\textit{W}}_f\right)^{-1}\textbf{\textit{W}}_f^\textbf{\textit{T}},\\
\end{equation}
where $(I/C+\textbf{\textit{W}}_f^T\textbf{\textit{W}}_f)^{-1}\textbf{\textit{W}}_f^\textbf{\textit{T}}$ is the MP inverse of $\textbf{\textit{W}}_f$. Then, the target change of other hidden layers is calculated by
\begin{equation}
\small
\label{n51}
\textbf{\textit{P}}^{(i-1)}=\sigma^{-1}\left({\textbf{\textit{P}}}^{(i)} \cdot \left(\frac{I}{C}+\left(\textbf{\textit{W}}_e^{(i)}\right)^T\textbf{\textit{W}}_e^{(i)}\right)^{-1}\cdot \left(\textbf{\textit{W}}_e^{(i)}\right)^T\right),\\
\end{equation}
where $2\leq i\leq M$, $\sigma^{-1}(\cdot)$ is the inverse of activation function, and $\textbf{\textit{P}}^{(i)}$ is the $i$-{th} layer target offset.

\subsubsection{Stage 3: Update Weights with MP Inverse}

In this paper, we hypothesize that the error $\textbf{\textit{E}}$ and feedback data $\textbf{\textit{P}}^{(i)}$ contain some information clues that AEs have not learned, and the weights optimized by the target offset $\textbf{\textit{P}}^{(i)}$ and encoded feature $\Psi^{(i)}_e$ can boost the feature encoding capacity and improve the generalization performance. After error backpropagation, at Stage 3, both the output layer and hidden layer weights of RML-MP are updated.

First, the hidden layer weights $\textbf{\textit{W}}_e^{(i)}$ are optimized. For the $i$-{th} hidden layer, an error-based update weight $\pmb{\eta}^{(i)}$ is calculated, satisfying
\begin{equation}
\small
\begin{split}
\label{n601}
&\hat{\Psi}_e^{(i-1)}(\textbf{\textit{W}}_e^{(i)}+\lambda\cdot\pmb{\eta}^{(i)})=\Psi_e^{(i)}+\textbf{\textit{P}}^{(i)}, \,\, 1\leq i\leq M\\
\end{split}
\end{equation}
where $\lambda$ stands for learning rate, $\hat{\Psi}_e^{(i-1)}$ is the updated $(i-1)$-{th} hidden layer representations. The $i$-{th} $(1\leq i\leq M)$ layer weight $\pmb{\eta}^{(i)}$ and $\hat{\Psi}_e^{(i)}$ are generated by using the MP inverse:
\begin{equation}
\small
\begin{split}
\label{n6}
\pmb{\eta}^{(i)}&=\left(\frac{I}{C}+(\hat{\Psi}_e^{(i-1)})^T\hat{\Psi}_e^{(i-1)}\right)^{-1}(\hat{\Psi}_e^{(i-1)})^T \textbf{\textbf{\textit{P}}}^{(i)},\\
\hat{\Psi}_e^{(i)}&=\begin{cases}\textbf{\textit{X}},$\,\,\,\,\,\,\,\,\,\,\,\,\,\,\,\,\,\,\,\,\,\,\,\,\,\,\,\,\,\,\,\,\,\,\,\,\,\,\,\,\,\,\,\,\,\,$ i=0\\\sigma\left(\hat{\Psi}_e^{(i-1)}\cdot \hat{\textbf{\textit{W}}}_e^{(i)}\right),$\,\,\,\,\,$1\leq i\leq M,\\\end{cases}
\end{split}
\end{equation}
where $\hat{\textbf{\textit{W}}}_e^{(i)}$ is the updated weight. When $i=1$, the weights of $1$-{st} hidden layer are updated. The input $\textbf{\textit{X}}$ is considered as the $0$-{th} layer feature $\hat{\Psi}_e^{(0)}$. Thus, the $i$-{th} hidden layer weight can be updated by
\begin{equation}
\small
\begin{split}
\label{n7}
\hat{\textbf{\textit{W}}}_e^{(i)}&=\textbf{\textit{W}}_e^{(i)}+\lambda\cdot \pmb{\eta}^{(i)}\\
&=\textbf{\textit{W}}_e^{(i)}+\lambda\cdot\left(\frac{I}{C}+(\hat{\Psi}_e^{(i-1)})^T\hat{\Psi}_e^{(i-1)}\right)^{-1}(\hat{\Psi}_e^{(i-1)})^T \textbf{\textbf{\textit{P}}}^{(i)},\\
\end{split}
\end{equation}

\noindent \textcolor{black}{Then, the weights of the output layer are updated as~\cite{kasun2013representational}}:
\begin{equation}
\small
\label{n8}
\hat{\textbf{\textit{W}}}_f=\left(\frac{I}{C}+(\hat{\Psi}_e^{(M)})^T\hat{\Psi}_e^{(M)}\right)^{-1}(\hat{\Psi}_e^{(M)})^T\cdot \textbf{\textit{T}},\\
\end{equation}

\subsubsection{{The Learning Steps of RML-MP}}

The proposed RML-MP can be summarized as follows.

\textbf{Step 1}: Given input features and the corresponding labels $\{\textbf{\textit{X}},\,\,\textbf{\textit{T}}\}$, the learning rate $\lambda$, and regularization term $C$.

\textbf{Step 2}: Train a multilayer LS-based network with $M$ AEs, analytically calculate the output layer parameters $\textbf{\textit{W}}_f$ with Eq.~(\ref{n3}).

\textbf{Step 3}: Pull back error term $\textbf{\textit{E}}$ from the classification layer to the first hidden layer by Eq. (\ref{n5}) and Eq. (\ref{n51}) one-by-one.

\textbf{Step 4}: Recalculate weights $\hat{\textbf{\textit{W}}}_e^{(i)}$ and $\hat{\Psi}_e^{(i)}$ from the $1$-{st} hidden layer to the $M$-{th} hidden layer via Eq.~(\ref{n6}) and Eq.~(\ref{n7}) sequentially.

\textbf{Step 5}: Update classification layer weights $\hat{\textbf{\textit{W}}}_f$ by Eq.~(\ref{n8}).

\subsection{The Proposed SRML-MP}

\subsubsection{Sparse Learning}

In SRML-MP, the updated weight $\pmb{\eta}^{(i)}$ for the $i$-th encoding layer is calculated as
\begin{equation}
\small
\label{n9}
{\rm min}\,\,J=\frac{1}{2}||\hat{\Psi}^{(i-1)}_e\pmb{\eta}^{(i)}-\textbf{\textit{P}}^{(i)}||^2_2+C||\pmb{\eta}^{(i)}||^{1/2}_{1/2},\\
\end{equation}
where $||\pmb{\eta}^{(i)}||^{1/2}_{1/2}=\sum_{j=1}^{N}|\pmb{\eta}_j^{(i)}|^{1/2}$. A fast iterative jumping thresholding (IJT)~\cite{zeng2016sparse} algorithm is adopted to solve Eq. (\ref{n9}). Specifically, with IJT~\cite{zeng2016sparse}, the proximity operator $pro_{\mu,C|\cdot|^q}$ of $\ell_q (0<q<1)$ can be described as
\begin{equation}
\small
\label{n10}
pro_{\mu,C|\cdot|^q}(z)=\begin{cases} \left(\cdot+\frac{C}{2}\mu q {\rm sign}(\cdot)^{q-1}\right)^{-1}(z),$\,\,\,$ |z|\geq \tau_{\mu,q}\\\,\,0,$\,\,\,\,\,\,\,\,\,\,\,\,\,\,\,\,\,\,\,\,\,\,\,\,\,\,\,\,\,\,\,\,\,\,\,\,\,\,\,\,\,\,\,\,\,\,\,\,\,\,\,\,\,\,\,\,\,\,\,\,\,\,\,\,\,$|z|\leq \tau_{\mu,q},\\\end{cases}
\end{equation}
for any $z\in R$, where
\begin{equation}
\small
\label{n11}
\begin{split}
&\tau_{\mu,q}=\frac{2-q}{2-2q}\left(C\mu(1-q)\right)^{\frac{1}{2-q}},\\
&\psi_{\mu,q}=\left(C\mu(1-q)\right)^{\frac{1}{2-q}}.
\end{split}
\end{equation}
The range of $pro_{\mu,C|\cdot|^q}$ is $\{0\}\cup [\psi_{\mu,q},\infty)$. \textcolor{black}{For $q=1/2$, with proximity operator, the $i$-th encoding layer weights $\pmb{\eta}^{(i)}$ can be expressed analytically~\cite{xu2012l_}:}
\begin{equation}
\small
\label{n12}
\pmb{\eta}^{(i)}=U_{\textbf{\textit{P}}^{(i)}}\cdot{\rm diag}\{\sigma_{\textbf{\textit{P}}^{(i)}}-\sqrt CP_{\ell1}(\frac{\sigma_{\textbf{\textit{P}}^{(i)}}}{\sqrt C})\}\cdot V_{\textbf{\textit{P}}^{(i)}}^T,\\
\end{equation}
\textcolor{black}{where $U_{\textbf{\textit{P}}}$, $V_{\textbf{\textit{P}}}$, and $\sigma_{\textbf{\textit{P}}^{(i)}}$ stand for the left unitary matrix of SVD of target output $\textbf{\textit{P}}^{(i)}$, the right unitary matrix of SVD of target output $\textbf{\textit{P}}^{(i)}$, and singular values of $\textbf{\textit{P}}^{(i)}$, respectively. $P_{\ell1}(\cdot)$ is the orthogonal projection of one vector onto the $\ell_1$ unit ball.}

\subsubsection{{The Learning Steps of SRML-MP}} The proposed SRML-MP is developed with four steps.

\textbf{Step 1}: Train a multilayer LS-based network with $M$ AEs, calculate output layer weights $\textbf{\textit{W}}_f$ via Eq.~(\ref{n3}).

\textbf{Step 2}: Pull back error $\textbf{\textit{E}}$ via Eq. (\ref{n5}) and Eq. (\ref{n51}).

\textbf{Step 3}: Calculate the $i$-{th} layer updated weight $\pmb{\eta}^i$ with $\ell_{1/2}$ penalty by IJT algorithm by Eq. (\ref{n12}), recompute the hidden layer weights via Eq. (\ref{n7}).

\textbf{Step 4}: Update the output layer weights $\hat{\textbf{\textit{W}}}_f$ with $\ell_{1/2}$ norm.

\section{Experiments}
\label{exp}
To efficiently evaluate the proposed RML-MP and SRML-MP with other RL algorithms, a set of experiments were first conducted in the traditional image classification domain. Experimental results on other real-world domains are listed in the supplementary file due to the space constraint.

\subsection{Experimental Setup}

All of the experiments conducted in this paper were performed in the Matlab R2019b environment, running in a workstation with Intel Core $E5-2650$ CPU and 256 GB memory. The high-level features extracted from DCNNs were carried out on the NVIDIA $1080Ti$ GPU. The Top-1 testing accuracy is adopted to evaluate different algorithms.

\subsubsection{{Datasets}}

In this paper, seven image classification datasets were used for experimentation: Caltech101, ImageNet-m/1, Place365-1/2/3, and Place365. The dataset specifications are shown in Table~\ref{dataset}. Details of each dataset are as follows.

\begin{table}
\small
\centering
\vspace{-0.2 cm}
\caption{Summary of the datasets.}
\vspace{0.2 cm}
\setlength\tabcolsep{9.0pt}
\begin{tabular}{llrrrcccccc}
\toprule
Datasets  &Classes &Training  &Testing  \\
\midrule
Caltech101    &102    &3060   &6084   \\
Place-1    & 365   & 146,000   & 36,500\\
Place-2          & 365  & 292,000   &73,000\\
Place-3          & 365  & 438,000   &109,500\\
ImageNet-1          & 1,000 & 400,000   &100,000\\
ImageNet-m          & 400  & 410,497   &102,623\\
Place365            & 365 & 1,803,460 &365,000\\
\bottomrule
\vspace{-0.4 cm}
\end{tabular}
\label{dataset}
\end{table}

The Place365 and ImageNet datasets are considered the largest datasets in the image classification area. The ImageNet set contains more than 1.2 million patterns. In this paper, two mini datasets are utilized. In particular, i) the ImageNet-1 dataset is generated by randomly selecting 500 images per category from the dataset, and ii) all the samples of the first 400 categories from ImageNet are collected to form the ImageNet-m dataset. The Place365 dataset is composed of 365 categories, containing more than $1.8 M$ images. 500, 1,000, and 1,500 images per category from this dataset are selected to generate the Place365-1, Place365-2, and Place365-3 datasets. For all of these (ImageNet-1/m, Place365-1/2/3) datasets, 80\% of the samples are adopted for training, whereas the rest are utilized for testing. Furthermore, the original Place365 dataset is used to validate the proposed RL algorithms, where the training set is applied for training and the validation set was for testing. Besides, Caltech101 is a widely used scene image classification dataset. Following the training settings in the previous works~\cite{zhang2020width, yang2019features}, we take 30 patterns per class for training and use the rest for testing.

It is noted that the traditional MP inverse utilizes a one-batch learning strategy to optimize the weights, which means the whole bunch of data is loaded and processed once. Thus, it needs industrial-scale computational resources (such as a workstation with 256 GB main memory) to handle large sets, like ImageNet. In fact, Zhang~\cite{zhang2020multimodel} proposed a batch-by-batch learning strategy for multilayer LS-based networks to process big datasets. The large data can be handled without occupying too much computational resources. Thus, the batch-by-batch strategy enables the proposed schemes to be processed on any computers.

\begin{figure}[!t]
\small
\centering
\includegraphics[trim={1.5cm, 0.0cm, 0.7cm, 1.0cm}, clip, width=0.5\textwidth]{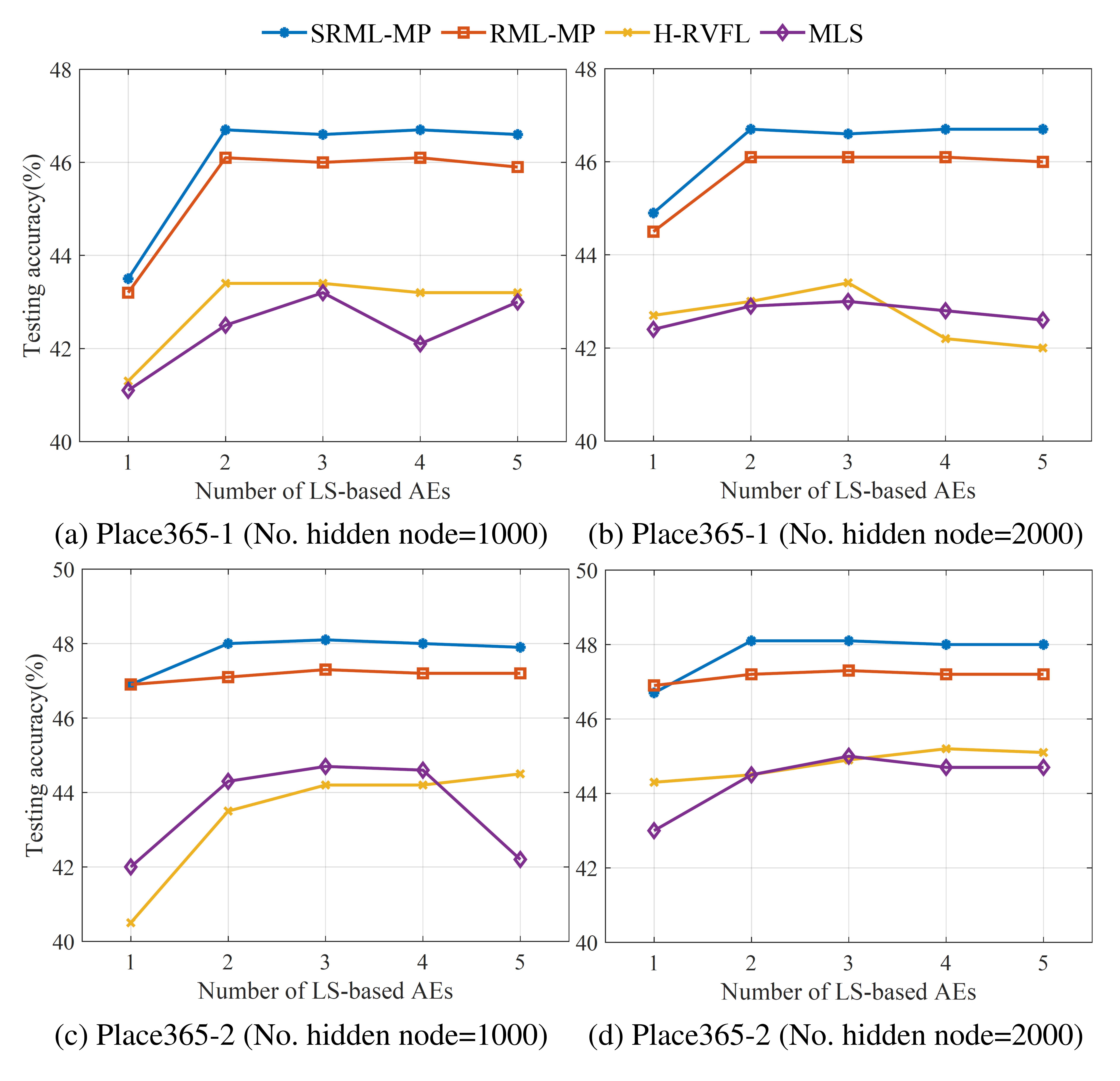}
\vspace{-0.9 cm}
\caption{\textcolor{black}{Comparison of different RL algorithms with Inception-v3 features. (a) and (b) are the results on Place365-1 dataset, (c) and (d) are the results on Place365-2 dataset.}}
\vspace{-0.2 cm}
\label{F3_com}
\end{figure}

\subsubsection{{Comparison to the State-of-the-Art}}

In this paper, several reconstruction-based RL algorithms are tested, which is divided into two families.

{RL algorithms with BP technique:} This family includes WD-AE~\cite{charte2018practical}, denoise autoencoder with Gaussian (DAE$^G$)~\cite{vincent2010stacked} and binary  (DAE$^B$)~\cite{vincent2010stacked} making noise, and SAE~\cite{ng2011sparse}. After the optimal encoding is learned via each RL algorithm, a softmax classifier is used to find the final classification result. \textcolor{black}{For the above-mentioned BP-based algorithms, stochastic gradient descent (SGD)-based fine-tuning is utilized to optimize the weights of the trained model.}

{RL algorithms with MP inverse technique:} These methods include hierarchical RVFL-based neural network (H-RVFL)~\cite{katuwal2019stacked}, multilayer LS-based framework (MLS)~\cite{kasun2013representational}, subnet-based structure (SNN)~\cite{wu2020multimodal}, hierarchical subnet framework (H-SNN)~\cite{yang2019features}, and width-growth model with subnetnode (W-SNN)~\cite{zhang2020width}. \textcolor{black}{Note that the weights of hidden layers are fixed after the unsupervised encoding is finished. In other words, these comparison algorithms do not contain any fine-tuning.}

\begin{table}[!t]
 \centering
 \small
 \vspace{-0.4 cm}
\caption{Effectiveness analysis of feature fusion (Vg. - VGG-16 features, In. - Inception-v3 features, Re. - ResNet50 features).}
\vspace{0.2 cm}
\setlength\tabcolsep{2.5pt}
\begin{tabular}{clccccccccc}
\toprule
Methods     &Features     &Place365-1     &Place365-3   &ImageNet-1\\
\midrule
\multirow{5}*{RML-MP}
&Vg.           &42.8   &44.4   &60.6\\
&Re.           &43.7   &47.1   &71.3\\
&In.           &46.0   &48.8   &81.2\\
&Re., In.      &47.0   &49.2   &81.4\\
&Vg., Re., In.  &48.2   &51.6   &\textbf{82.0}\\
\midrule
\multirow{5}*{SRML-MP}
&Vg.            &43.7   &44.6   &60.1\\
&Re.           &44.5   &47.0   &70.9\\
&In.           &46.7   &49.1   &81.2\\
&Re., In.      &47.9   &50.6   &81.2\\
&Vg., Re., In.  &\textbf{49.7}   &\textbf{52.1}   &81.8\\
\bottomrule
\vspace{-0.6 cm}
\label{t1_n}
\end{tabular}
\end{table}

For BP-based RL strategies, the training epochs and size of each mini-batch are set to be 100. The initial learning rate is set as 0.01 with a 0.1 attenuation rate for every 10 training epochs. The input corruption rate of DAE$^G$ is 0.5. \textcolor{black}{The training epochs for fine-tuning is 20.} For MP inverse-based RL algorithms, the number of hidden neurons and the regularization term in all AEs are optimized within the grid $\{500,\,\,1000,\,\,2000\}\times\{10^{-3},\,\,10^0,\,\,10^3\}$, while the regularization term for the last classification layer is searched from $\{10^0,\,\,10^2,\,\,10^4\}$. The number of subnets is set as 5, and the number of neurons in each subnet is defined as 1000. Also, the regularization term in subnet-based models is optimized within the grid $\{10^0,\,\,10^2,\,\,10^4\}$. \textcolor{black}{As for the proposed RML-MP and SRML-MP, the number of stacked AEs is 2, and the number of hidden neurons for each LS-based AE is 1000. The sigmoid function is chosen as the non-linear activation function. The offset term $C$ is 4.}

\begin{table*}[!t]
\small
 \centering
 \vspace{-0.2 cm}
\caption{Top-1 testing accuracy comparison among different non-iterative RL methods. Values in \textcolor{black}{BLUE} are the best results with Inception-v3 features. The ones in \textcolor{red}{RED} are the best results with cancatenated features (Inc. - Inception-v3 features).}
 \vspace{0.2 cm}
\setlength\tabcolsep{6.0pt}
\begin{tabular}{lcccccccccc}
         \toprule
        Methods    &Caltech101    &Place365-1     &Place365-2     &Place365-3     &ImageNet-1    &ImageNet-m   &Place365    &Average\\
         \midrule
         \multicolumn{9}{c}{\emph{\textbf{Single-Model: Inception-v3 Features}}}\\
         \midrule
Inc.+H-RVFL  &90.2   &43.4   &45.2   &46.1   &78.6   &83.5   &48.8   &62.3\\
Inc.+H-SNN &89.4   &44.9   &46.4   &47.2   &79.8   &85.1   &48.5   &63.0\\
Inc.+MLS &89.7   &43.2   &45.0   &46.3   &76.5   &81.9   &46.8   &61.4\\
Inc.+SNN    &88.6   &42.5   &45.9   &46.8   &79.0   &84.5   &48.1   &62.2\\
Inc.+W-SNN   &89.6   &44.9   &46.8   &47.6   &79.6   &85.2   &49.7   &63.3\\
Inc.+RML-MP                       &91.7   &46.0   &47.3   &48.8   &\textcolor{blue}{81.3}   &86.2   &\textcolor{blue}{52.5}   &64.7\\
Inc.+SRML-MP                      &\textcolor{blue}{91.8}   &\textcolor{blue}{46.7}   &\textcolor{blue}{48.0}   &\textcolor{blue}{49.1}   &81.2   &\textcolor{blue}{86.4}   &52.1   &\textcolor{blue}{65.1}\\
         \midrule
         \multicolumn{9}{c}{\emph{\textbf{Multi-Model: Concatenated Features}}}\\
         \midrule
All+H-RVFL   &91.4   &46.3   &48.5   &48.5   &80.0   &85.2   &51.4    &64.5\\
All+H-SNN  &92.2   &47.1   &48.6   &49.3   &\textcolor{red}{82.5}   &86.6   &51.6   &65.4\\
All+MLS  &91.7   &46.6   &47.4   &48.8   &80.2   &84.3   &50.0   &64.1\\
All+SNN~     &91.8   &45.3   &46.7   &47.4   &79.9   &82.1   &50.4   &63.4\\
All+W-SNN    &92.1   &47.1   &48.9   &50.2   &81.3   &87.0   &52.3   &65.6\\
All+RML-MP                        &93.2   &48.2   &51.0   &51.6   &82.0   &\textcolor{red}{87.2}   &\textcolor{red}{54.8}   &66.8\\
All+SRML-MP                       &\textcolor{red}{93.4}   &\textcolor{red}{49.7}   &\textcolor{red}{51.2}   &\textcolor{red}{52.1}   &81.8   &87.1   &54.1    &\textcolor{red}{67.1}\\
         \bottomrule
\label{t2_n}
\end{tabular}
\end{table*}

\subsubsection{{Multi-Model Features}}

Multi-model algorithms fuse different sources of features that are complementary to each other to achieve superior recognition performance. Three pre-trained DCNNs, namely VGG-16~\cite{szegedy2015going}, ResNet-50~\cite{he2016deep}, and InceptionNet-v3~\cite{szegedy2016rethinking}, are adopted as feature extractors. The above-mentioned DCNNs are pre-trained on the ImageNet set, and the final softmax layer contains 1,000 neurons. In this paper, the high-level features that are extracted from the top layer of DCNNs are loaded as the raw feature. As for high-level feature extraction, the original top layer is decapitated and replaced with a new classification layer (and softmax layer) with the same number of classes as the target dataset. After the DCNN is fine-tuned, the high-level features are captured from the penultimate layer of each DCNN. In particular, the data extracted from VGG-16, ResNet-50, and InceptionNet-v3 are 4,096-, 2,048-, and 2,048-dimensional vectors, respectively. Thus, the dimension of the concatenated feature is 8,192.

\begin{figure*}[!t]
\centering
\includegraphics[trim={1.0cm, 0.0cm, 1.0cm, 1.0cm}, clip, width=\textwidth]{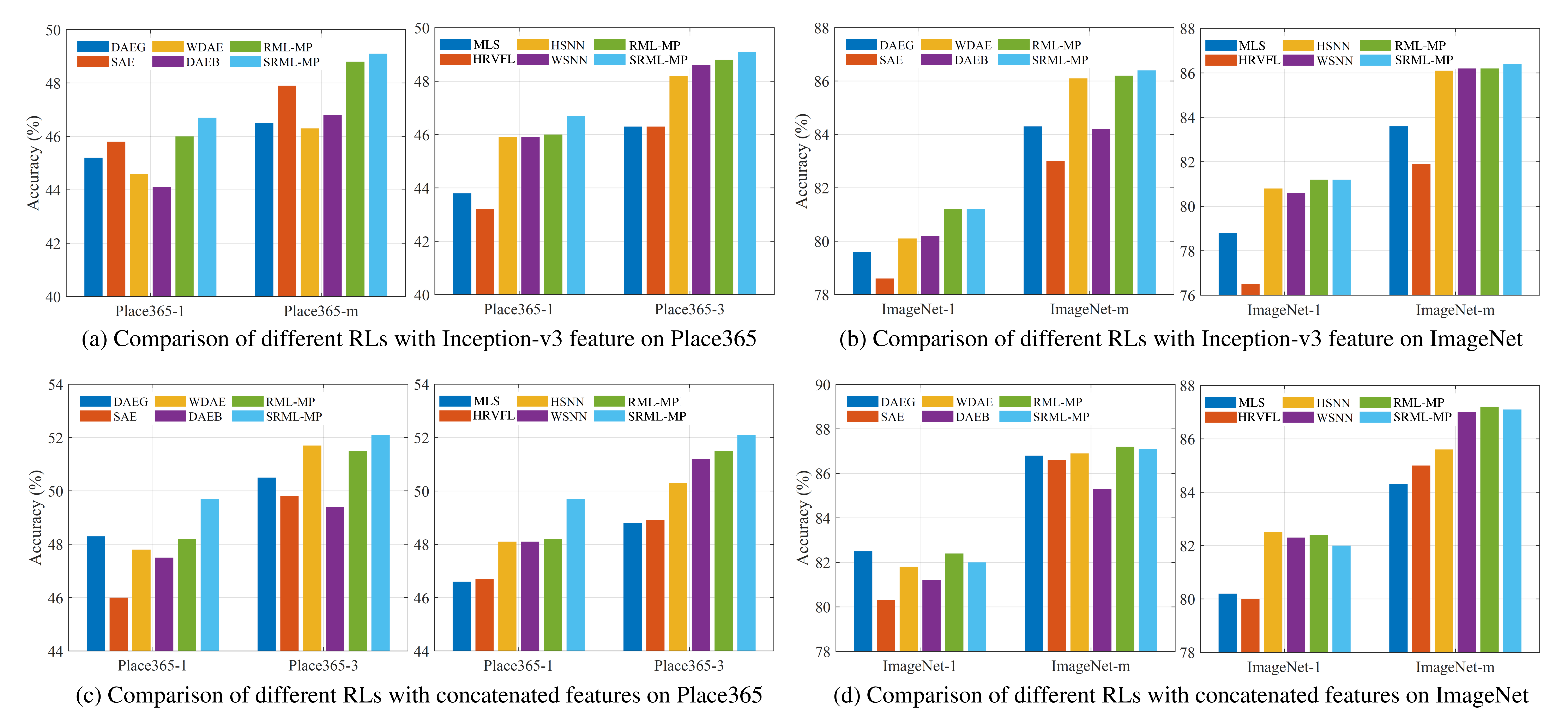}
\vspace{-0.9 cm}
\caption{\textcolor{black}{Comparison of different reconstruction-based RL algorithms on Place365 and ImageNet datasets. (a) and (b) are the comparisons with Inception-v3 feature, (c) and (d) are the results with all concatenation feature.}}
\label{F1:asa}
\end{figure*}

\subsection{Quantitative Analysis}

\subsubsection{\textcolor{black}{{Model Settings}}}

The proposed algorithms have several hyperparameters, such as the number of AEs used, the number of neurons in each AE. In this subsection, we empirically verify the recommendations of these two hyperparameters. \textcolor{black}{Figure~\ref{F3_com} shows the Top-1 testing accuracy of the proposed methods, MLS, and H-RVFL on Place365-1/2 datasets as the number of AEs increased. In this evaluation, although each algorithm's learning procedures differ, the offset term $C$ and the number of hidden neurons in each AE remain the same (1000 or 2000) for consistency and fair comparison. Fig.~\ref{F3_com} (a) and (c) depict the results when the number of hidden neurons in each AE equals 1,000, while Fig.~\ref{F3_com} (b) and (d) show the comparison results when the number of latent space nodes in each AE is 2000. It is observed that i) the performance of RML-MP and SRML-MP converge after the number of LS-based AEs equals 2, ii) the proposed models achieve competent performance when the number of neurons in each AE is 1000, and iii) with the same hyperparameters, the Top-1 testing accuracy of the proposed models is improved by 2\% when compared to the existing multilayer LS-based algorithms, such as H-RVFL and MLS. Thus, in this paper, for the proposed RML-MP and SRML-MP, the number of stacked AEs is set to 2, and the number of hidden neurons in each AE is 1000.}



\subsubsection{{Experiments on Image Classification}}


To verify the effectiveness of multi-model data learning, a sanity check with various combinations of high-level features is conducted as reported in Table~\ref{t1_n}. Different combinations from single-model to multi-model features are loaded as input. The experimental results show that the fusion of multi-model features provides a remarkable performance boost over models trained with single-model features. For example, the proposed RML-MP with the concatenated features (ResNet + InceptionNet + VGG) provides 48.2\%, 51.6\%, and 82.0\% testing accuracy on the Place365-1, Place365-3, and ImageNet datasets, respectively, having 5.4\%, 7.2\%, and 21.4\% higher Top-1 testing accuracy than that of the single-model VGG feature.

The overall comparisons of the existing multilayer LS-based RL algorithms and the proposed methods on the image classification datasets are provided in Table~\ref{t2_n}. Along with the Top-1 testing accuracy, the mean average performance among all datasets is shown. Through comparison, the following conclusions can be drawn: i) The proposed RML-MP and SRML-MP with single-model feature provide superior performance compared to the comparison algorithms. For instance, the SRML-MP shows a valuable increment over the MLS, including 2.8\% on Place365-3, 4.7\% on ImageNet-m, and 5.7\% on Place365, respectively. ii) The RML-MP and SRML-MP with the concatenated features have competent performance over other RL frameworks. For example, the mean average accuracy of RML-MP and SRML-MP is promoted by 2.7\% and 3.0\% than that of the MLS, respectively. iii) Compared to SRML-MP, RML-MP offers benefits in processing big and complex datasets (with more than 1 million samples). For example, when only considering multi-model concatenated feature, RML-MP achieves 0.7\% better accuracy than the SRML-MP on Place365 dataset.

\begin{table}[!t]
\small
 \centering
 \vspace{-0.2 cm}
\caption{Processing time with Inception-v3 features in minute.}
\vspace{0.2 cm}
\setlength\tabcolsep{6pt}
\begin{tabular}{lrrrrrrrrrrrrr}
\toprule
Methods     &Place-1     &Place-2   &Place-3       &ImageNet-1\\
\midrule
\multicolumn{5}{c}{\emph{\textbf{Training Time Comparison}}}\\
\midrule
MLS  &1.5    &3.0    &5.9    &5.9\\
H-RVFL   &2.0    &3.9    &9.5    &8.5\\
SNN   &14.7   &30.9   &56.7   &54.9\\
H-SNN   &15.2   &31.4   &59.3   &57.8\\
RML-MP &4.2    &9.4    &21.5   &20.2\\
SRML-MP&5.6    &11.8   &28.2   &26.6\\
\midrule
\multicolumn{5}{c}{\emph{\textbf{Testing Time Comparison}}}\\
\midrule
MLS  &0.1    &0.2    &0.9    &0.9\\
H-RVFL   &0.2    &0.4    &1.3    &1.3\\
SNN   &0.2    &0.4    &1.7    &1.7\\
H-SNN   &0.2    &0.5    &1.9    &2.0\\
RML-MP &0.1    &0.2    &0.9    &0.9\\
SRML-MP&0.1    &0.2    &0.9    &0.9\\
\bottomrule
\label{t3_n}
\end{tabular}
\vspace{-0.5 cm}
\end{table}

Furthermore, the comparison results with all state-of-the-art reconstruction-based RL algorithms are shown in Fig.~\ref{F1:asa}. As can be observed in the figure, the proposed RML-MP and SRML-MP generally outperform the rest of the reconstruction-based RL algorithms. Therefore, the benefits of the proposed RML-MP and SRML-MP on big data analysis are verified.

\subsubsection{{Computational Time Analysis}}

The processing time of different RL algorithms is reported in this subsection. Table~\ref{t3_n} tabulates the training and inference time complexities of the existing LS-based methods and the proposed RML-MP and SRML-MP. All of the results recorded in Table~\ref{t3_n} are counted in minutes. For a fair comparison, the DCNN training time and feature pre-processing time are ignored. Only the complexities of network encoding and pattern classification are recorded. One can readily see from the table that the MLS needs the shortest training time, while the proposed RML-MP and SRML-MP take around two to three times longer for training. The reason for the longer training time is that, on one dataset, the proposed learning pipelines search the better representations by pulling back the error matrix from the final classification layer to each hidden layer respectively. Although the proposed MRL-MP and SMPL-MP need longer training time, they are more effective as they provide much better generalization performance than the existing multilayer methods.

As for the inference time, the RML-MP and SRML-MP provide similar inference time on four datasets as that of MLS. Other RL algorithms such as H-RVFL, W-SNN, and H-SNN roughly demand 30\%, 90\%, and 110\% more inference time than RML-MP and SRML-MP, respectively.

\section{Conclusion}
\label{con}

The paper proposes two multilayer neural networks for multi-model large data analysis. The RML-MP and SRML-MP are developed to enhance the generalization capability of the traditional multilayer LS-based structures. The contributions of this paper are as follows: First, representations learned from traditional least squares-based autoencoders may be biased and inadequate for solving complex tasks (such as ImageNet). In this paper, the retraining strategy is proposed to enhance the representation capacity. Second, the proposed RML-MP and SRML-MP handle big data efficiently. These models process large-scale datasets such as Place365 containing more than 1.8 million samples with competitive performance and affordable training time. The experiments on 7 datasets show that the RML-MP and SRML-MP provide superior performance to the existing multilayer LS-based representation learning algorithms.

\nocite{langley00}

\bibliography{example_paper}
\bibliographystyle{icml2021}

\end{document}